\documentclass[journal,twoside,web]{ieeecolor}
\usepackage{tmi}
\usepackage{cite}
\usepackage{amsmath,amssymb,amsfonts}
\usepackage{algorithmicx}
\usepackage{algpseudocode}
\usepackage{algorithm}
\usepackage{makecell}
\usepackage{array}
\usepackage{graphicx}
\usepackage{textcomp}
\usepackage{caption}
\usepackage{siunitx}
\usepackage[export]{adjustbox} 
\usepackage{multirow}
\usepackage[symbol]{footmisc}

\def\BibTeX{{\rm B\kern-.05em{\sc i\kern-.025em b}\kern-.08em
    T\kern-.1667em\lower.7ex\hbox{E}\kern-.125emX}}
\begin{document}
\title{Which Pixel to Annotate: a Label-Efficient Nuclei Segmentation Framework}
\author{Wei Lou, Haofeng Li, Guanbin Li, Xiaoguang Han, and Xiang Wan
\thanks{Wei Lou and Haofeng Li contributed equally to this work. Corresponding author: Haofeng Li. }
\thanks{Wei Lou, Haofeng Li and Xiaoguang Han are with Shenzhen Research Institute of Big Data, Guangdong Provincial Key Laboratory of Big Data Computing, The Chinese University of Hong Kong at Shenzhen, Shenzhen 518172, China (e-mail: 221019047@link.cuhk.edu.cn; lhaof@sribd.cn; hanxiaoguang@cuhk.edu.cn).}
\thanks{Guanbin Li is with the School of  Computer Science and Engineering, Sun Yat-sen University, Guangzhou 510006, China (e-mail: liguanbin@mail.sysu.edu.cn).}
\thanks{Xiang Wan is with Shenzhen Research Institute of Big Data, Guangdong Provincial Key Laboratory of Big Data Computing, The Chinese University of Hong Kong at Shenzhen, Shenzhen 518172, China, and also with Pazhou Lab, Guangzhou, 510330, China (wanxiang@sribd.cn).}
\thanks{This work is supported by Chinese Key-Area Research and Development Program of Guangdong Province (2020B0101350001), in part by the National Natural Science Foundation of China (No.62102267), in part by NSFC under the project “The Essential Algorithms and Technologies for Standardized Analytics of Clinical Texts” (12026610), in part by the Guangdong Basic and Applied Basic Research Foundation (No.2020B1515020048), in part by Open Research Projects of Zhejiang Lab (No.2019KD0AD01/017), in part by the National Natural Science Foundation of China (No.61976250), in part by NSFC-61931024 and Shenzhen Sustainable Development Project(KCXFZ20201221173008022) and the Guangdong Provincial Key Laboratory of Big Data Computing, The Chinese University of Hong Kong, Shenzhen.}
}

\maketitle

\begin{abstract}
Recently deep neural networks, which require a large amount of annotated samples, have been widely applied in nuclei instance segmentation of H\&E stained pathology images. However, it is inefficient and unnecessary to label all pixels for a dataset of nuclei images which usually contain similar and redundant patterns. Although unsupervised and semi-supervised learning methods have been studied for nuclei segmentation, very few works have delved into the selective labeling of samples to reduce the workload of annotation. Thus, in this paper, we propose a novel full nuclei segmentation framework that chooses only a few image patches to be annotated, augments the training set from the selected samples, and achieves nuclei segmentation in a semi-supervised manner. In the proposed framework, we first develop a novel consistency-based patch selection method to determine which image patches are the most beneficial to the training. Then we introduce a conditional single-image GAN with a component-wise discriminator, to synthesize more training samples. Lastly, our proposed framework trains an existing segmentation model with the above augmented samples. The experimental results show that our proposed method could obtain the same-level performance as a fully-supervised baseline by annotating less than 5\% pixels on some benchmarks.
\end{abstract}


\begin{figure}[ht]
\includegraphics[width=1.0\linewidth]{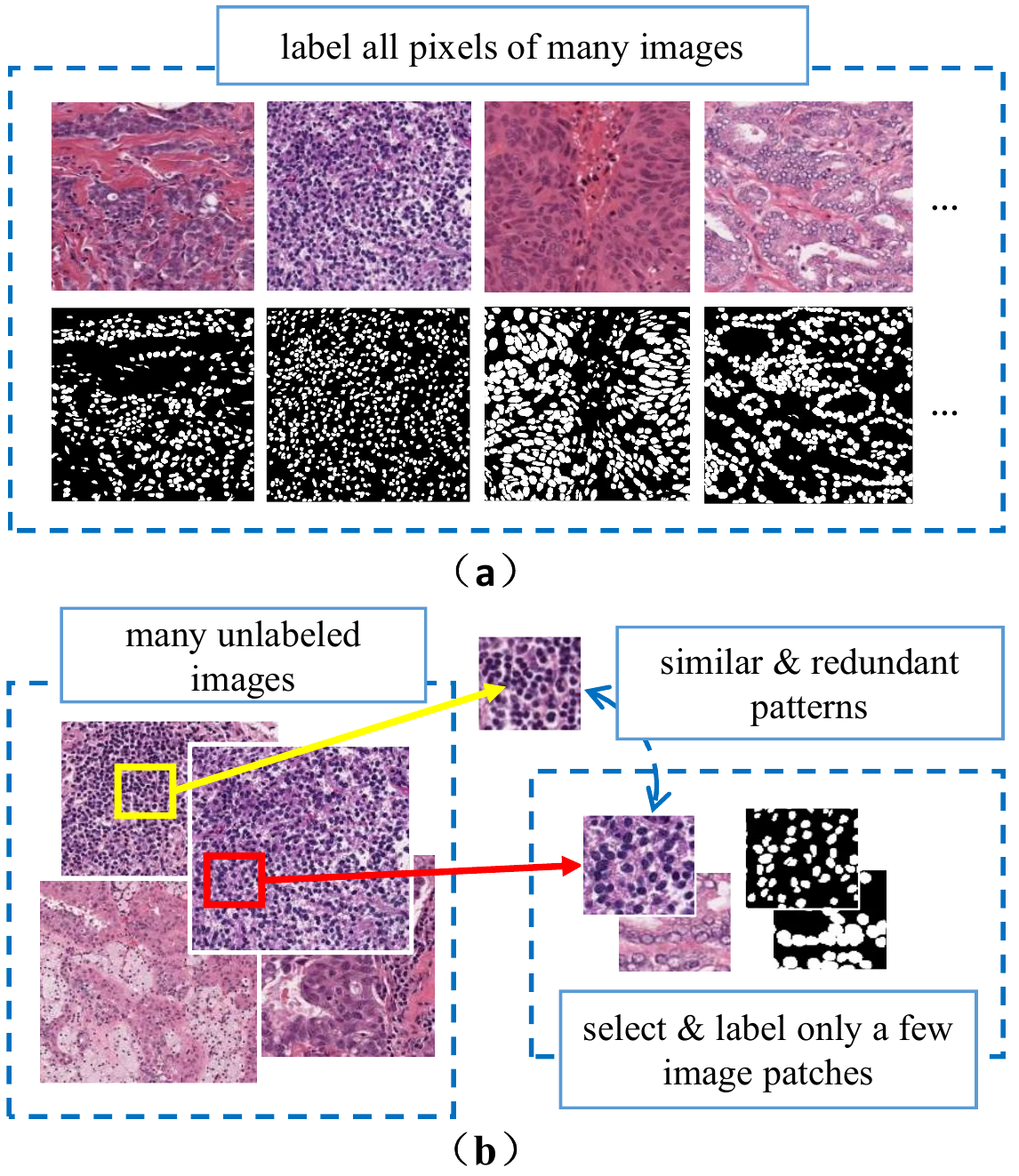}
\centering
\caption{Idea of image patch selection in the proposed nuclei segmentation framework. (a). Standard nuclei segmentation algorithms usually annotate all pixels of many nuclei images. (b). The proposed method only samples a few small image patches for labeling. Even with a smaller number of annotations, our proposed label-efficient framework achieves the same-level segmentation performance as a fully supervised baseline. }
\label{fig_idea}
\end{figure}

\begin{figure*}[h]
\captionsetup{justification=centering}
\includegraphics[width=0.9\linewidth]{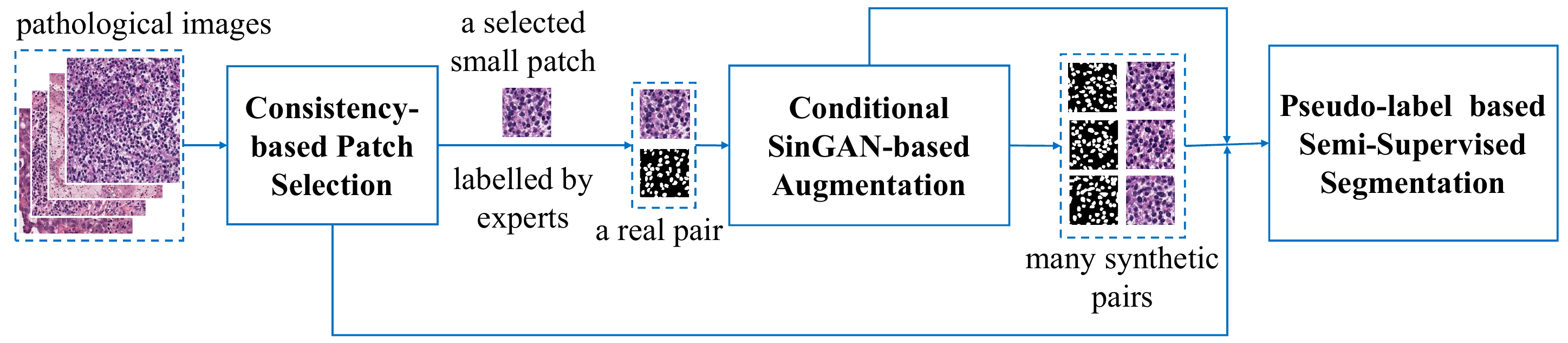}
\centering
\caption{Overview of our proposed nuclei segmentation framework.}
\label{fig_framework}
\end{figure*}

\section{Introduction}
\label{sec:introduction}
\IEEEPARstart 
N{\lowercase{u}}clei segmentation aims at labeling all pixels for each single nucleus in a histopathology image. The task could provide fundamental visual information and morphological features including the size, shape or color of nuclei~\cite{Clayton1991Pathologic,Elston1991Pathologic}, which are not only beneficial for middle-level understanding of histopathology image (for examples, cell classification, tissue segmentation, etc), but also related to high-level clinical analysis such as cancer diagnosis, assessment and prognostic prediction. Thus nuclei segmentation acts as a critical prerequisite in computer-aided diagnosis systems (CAD)~\cite{Xing2016Robust}. Due to the distraction by complex background, the clutter and partial occlusion of nuclei, separating each nucleus accurately still remains a challenging problem.

Currently, fully supervised models~\cite{liu2021panoptic,Kumar2017adataset,Naylor2018Segmentation,Chen2017DCAN,Graham2019Hover} are the most popular nuclei segmentation methods for their high accuracy, but they need pixel-level annotations which are expensive and time-consuming. Unsupervised methods based on domain adaptation~\cite{Liu2020Unsupervised,Tzeng2017Adversarial,Hsu_2021_CVPR} adapt an unlabeled dataset to a labeled one, while in real applications it is acceptable to annotate some small image patches. 
Semi-supervised segmentation methods~\cite{Li2020Self,Qu2020Weakly,Socher2010Connecting,Zhou2019Collaborative} improve the performance by training with both labeled and unlabeled images, but they rarely discuss the impact of selecting different samples for labeling on performance. Active learning-based methods \cite{Yang2017Suggestive,Zhou2017Fine-tuning} choose high-quality samples for labeling in an iterative way. They rely on an iteratively-trained model and few of them investigate the sample selection when no annotations are available. Since histopathology images consist of similar textures and redundant patterns, there is no need to label the entire image. As shown in Fig.~\ref{fig_idea}, we claim that searching for several representative small patches from the whole dataset may be sufficient for training a segmentation model.

After locating and labeling several image patches, how to exploit such a small amount of data for training poses another challenge. A straightforward idea is to synthesize more training pairs from the existing ones. In recent years generative adversarial networks (GANs)~\cite{Ian2014Generative,Arjovsky2017Wasserstein,Karras2019a}, which adversarially train a discriminator to help improve the generator, have achieved success in sampling images from noise vectors. Conditional GANs~\cite{Mirza2014Conditional,Reed2016Generative,Isola2017Image,Zhu2017Unpaired} could generate images from given priors such as labels, masks or texts. The above methods require collecting a number of samples to avoid over-fitting. Some recent work~\cite{Shaham2019Singan} can utilize only a single image for training but it is unconditional. Thus, it is worthwhile to conceive a conditional GAN that could efficiently augment a lot of training pairs from a single one. 

Motivated by the above observations, we propose a novel framework that learns to segment nuclei in a label-efficient way, as shown in Fig.~\ref{fig_framework}. The proposed method consists of sample selection, data augmentation and instance segmentation. To select the useful image patches for labeling, we enumerate a wide range of possible image patches and find out those with high representativeness and intra-consistency. We experimentally verify that such selected samples could better represent the whole training data and ease the GAN training at a later stage. Since only labeling a small amount of patches is not enough for training an effective segmentation model, we develop an efficient conditional single-image GAN that can be trained with only one pair of image and mask. Afterwards, we train an existing nuclei segmentation model with these augmented samples in a semi-supervised manner. 

To summarize, our contributions have four folds:
\begin{itemize}
\item We propose a novel label-efficient nuclei segmentation framework for semi-supervised setting.
\item  We develop a consistency-based patch selection method that only selects a part of image patches to annotate and reduces the labeling cost significantly.
\item We propose a data augmentation mechanism based on a conditional SinGAN with a novel component-wise discriminator.
\item The experiments show that our proposed framework could attain the approximate performance with a fully-supervised baseline by labeling less than 5\% pixels on some datasets.
\end{itemize}

\section{Related Work}
\subsection{Deep Learning-based Nuclei Segmentation}
Nuclei segmentation is a critical step before the clinical analysis in modern computer-aided diagnosis systems. The goal is to locate the boundaries of nuclei in histopathology images. Fully supervised methods \cite{liu2021panoptic,Kumar2017adataset,Naylor2018Segmentation,Chen2017DCAN,Graham2019Hover,Zhou2019Cia-net,zhao2020triple,chen2020boundary,xie2020instance,wang2020bending,qu2020nuclei,yi2021object} use a large-scale labeled dataset to train a segmentation model. Among them, Mask-RCNN~\cite{He2017Mask} is a popular object segmentation method for natural and medical images. It is a two-stage segmentation model that predicts bounding boxes for nuclei and then segments them inside the predicted boxes. Hover-net \cite{Graham2019Hover}, which is a unified FCN model for segmentation and classification of kernel instances, effectively encodes the horizontal and vertical distance information from kernel pixels to its centroid. However, most of these methods require a large number of labeled images which increase tremendous time and money cost for data collection and annotation. We adapt these two segmentation baselines with our label-efficient nuclei segmentation framework to exhibit the generalization of our methods.

Unsupervised methods~\cite{Liu2020Unsupervised,Sahasrabudhe2020Self,Hsu_2021_CVPR} are proposed to segment nuclei of a new dataset without annotations. Some are targeting the translation from a labeled domain to an unlabeled domain. PDAM~\cite{Liu2020Unsupervised} uses a semantic segmentation branch with a domain discriminator to narrow the domain gap between an unlabeled histopathology dataset and a labeled microscopy dataset. DARCNN~\cite{Hsu_2021_CVPR} adapts a labeled dataset of natural images to unlabeled datasets of biomedical images. However, these domain adaption based methods need to collect a large annotated dataset as the source domain in advance. Sahasrabudhe~{\it{et al.}}~\cite{Sahasrabudhe2020Self} adopt the prediction of magnification level as a self-supervision, and take an intermediate attention map of the neural model as the unsupervised result of nuclei segmentation. In real applications, it is usually affordable to annotate some image patches for a new dataset. Thus in this paper we adopt semi-supervised setting which is more practical.

\subsection{Semi-Supervised Learning}
Semi-supervised learning \cite{Li2020Self,Qu2020Weakly,Socher2010Connecting,Zhou2019Collaborative,bortsova2019semi,jeong2019consistency,zhou2021ssmd} is a common scenario in medical applications, where only a small subset of the training images is assumed to have full annotations. Among existing semi-supervised methods, we only briefly introduce pseudo labeling~\cite{lee2013pseudo,arazo2020pseudo,zhang2020label,Shi2018ECCV} which is most related to our work. In pseudo-label methods, a small amount of labeled data is first used to train a target model, and then the trained model is used to produce pseudo labels for unlabeled data. The pairs of pseudo labels and their corresponding images are added to the training set for the next round of model training. Although we use a semi-supervised setting in this paper, we mainly focus on selective labeling which is seldom considered by existing semi-supervised methods. We simply employ a vanilla pseudo-labeling method in our framework, which is to predict pseudo masks for unlabeled image patches using a pre-trained segmentation model trained by a few labeled patches. 

\subsection{Sample Selection}
The sample selection~\cite{bellver2020mask,mahapatra2018efficient,an2016semi} aims to select the most useful samples from an unlabeled dataset and label them as training data to avoid annotating the whole dataset. Active learning is a commonly used sample selection method. Active learning based methods~\cite{dagan1995committee,krishnamurthy2002algorithms,Yang2017Suggestive,Zhou2017Fine-tuning} reduce the annotation cost by iteratively labelling the high-value samples selected by a continuous training model. First, the training model is pre-trained or randomly initialized, and the unlabeled images are predicted by the training model and sorted according to uncertainty strategy. Second, experts are required to label the samples with high uncertainty. Then, these labeled samples are used to fine-tune and update the training model. Finally, repeat these steps until the performance converges or the available annotation resources (money, experts) are exhausted. However, on one hand, active learning methods seldom discuss which samples to choose at the very beginning when no trained models are available. On the other hand, very few researchers focus on the study of which regions in pathological images are more effective for nuclei segmentation learning than other regions. Meanwhile, the iterative learning and annotation process can still be costly. Therefore, our sample selection method focuses on the selection of small areas inside large images without iterative training.

\subsection{GAN-based Image Synthesis}
Image synthesis is the process that produces synthetic images in the target image domain (named as target domain) from the input images that reside in a different image domain (named as source domain). This process has been widely used for data augmentation. Generative Adversarial Networks (GANs) \cite{Ian2014Generative,Arjovsky2017Wasserstein,Karras2019a,majurski2019cell,kwon2019generation,yang2018low,yang2017dagan,li2018context} have made great success in image synthesis. GANs learn a discriminator loss to classify if the output image is real or fake and train the generator to minimise the error. Traditional GANs (unconditional GANs) \cite{Ian2014Generative,Arjovsky2017Wasserstein}  synthesize images from random noise. Since the generation process is trained without restrictions, the synthesized images may not have certain properties (\textit{e.g.}, corresponding to some given mask) that we need.

Therefore, the approaches using conditional image constraints to produce images with the desired properties have been proposed. Conditional-GANs (cGANs) \cite{Mirza2014Conditional,Reed2016Generative} synthesize images with prior information (conditions) which can be labels, masks or texts. Isola~\textit{et al.} \cite{Isola2017Image} employ many paired image-mask data to train a generative model to solve label-to-image tasks. Zhu~\textit{et al.} \cite{Zhu2017Unpaired} introduce cycleGANs that use many unpaired images from two different domains to train a domain-translation generator by minimizing a cycle-consistency loss. However, some of these methods require a large number of training images pairs that cause unaffordable costs. Shaham \textit{et al.} \cite{Shaham2019Singan} proposed an unconditional single-image GAN (sinGAN) which can efficiently sample images from random noises by training with only a single real image. In this paper we propose a novel conditional single-image GAN that achieves the translation from mask to image using only a single real pair of nuclei image and mask.

\begin{figure*}[!htb]
\includegraphics[width=1.0\linewidth]{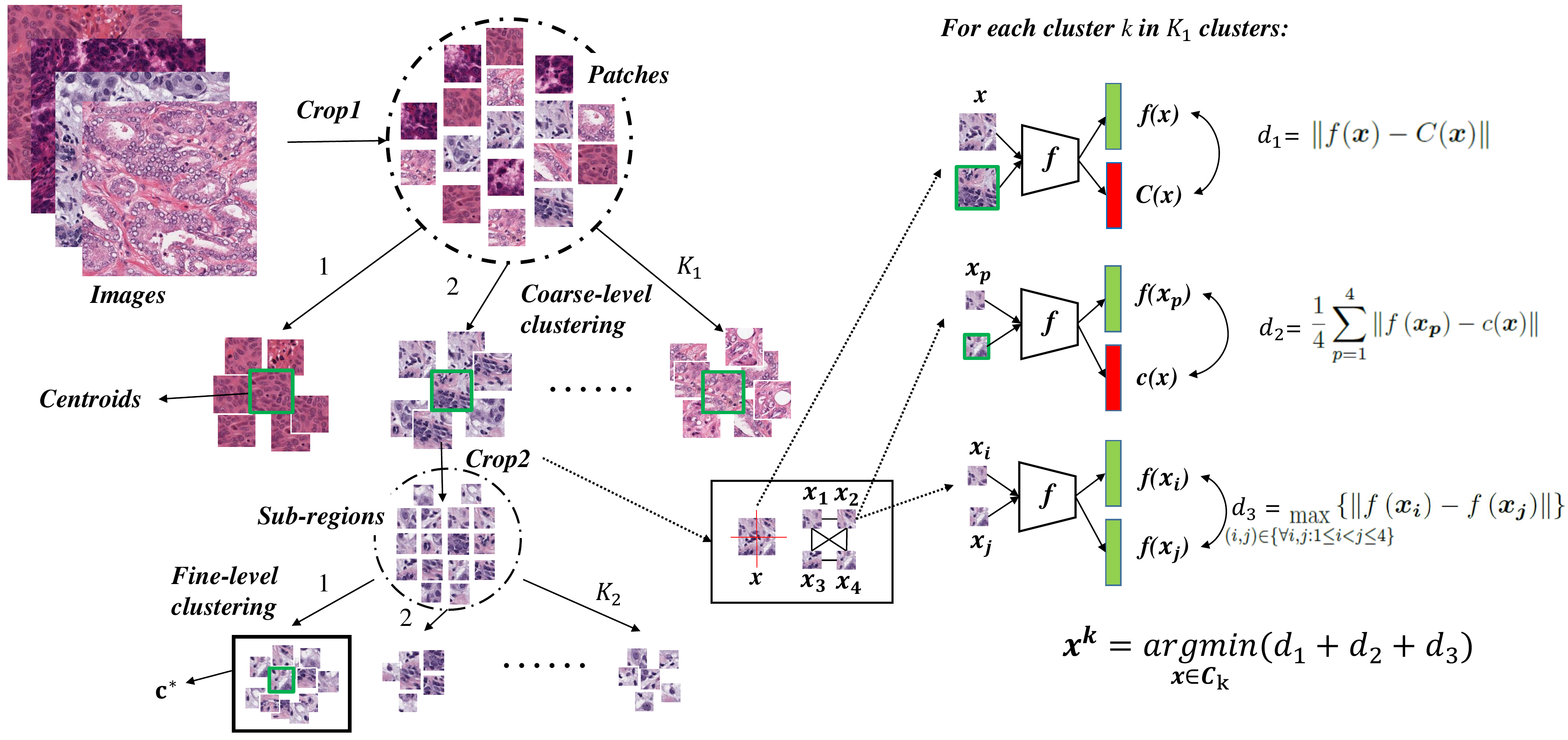}
\centering
\caption{Consistency-based Patch Selection (CPS). The left half shows the coarse-level and fine-level clustering process. The image patches/sub-regions with green borders are the one closest to their cluster center. The right half shows the criteria to select an image patch. `Crop1' is to cut an image into $s \times s$ image patches. `Crop2' is to cut an image patch into $\frac{s}{2} \times \frac{s}{2}$ sub-regions. $\boldsymbol{c^*}$ denotes the largest cluster among $K_2$ fine-level clusters. $\boldsymbol{x^k}$ denotes the selected image patch from the $k$-th cluster among $K_1$ coarse-level clusters.}
\label{fig_cps}
\end{figure*}

\section{Method}
In this section, we describe the proposed label-efficient nuclei segmentation framework from three aspects: how to determine the image patches to be annotated, how to synthesize more training pairs with those labeled patches, how to train a segmentation model using both labeled and unlabeled images.

\subsection{Consistency-based Patch Selection}\label{sec_CPS}
To locate the image patches that mostly benefit the final performance of nuclei segmentation, we consider two attributes of an image patch, an inter-patch attribute called {\it representativeness}, and an intra-patch one called {\it consistency}. Representativeness is an important inter-patch attribute that reflects the relationship among different image patches in a large dataset. Consider that in some latent space, the image patches that are relatively close to each other form a cluster. If the distance between some image patch $x$ and the others in a cluster is the smallest, then $x$ is better representative than any other members in the cluster. Hence, higher representativeness means smaller distance from the cluster center. We propose to label the image patches representing different clusters so that these image patches can well replace the whole trainset and decrease the redundancy.

To ease the GAN training for later data augmentation, we suggest to select the image patches with high intra-patch consistency, which measures the level of self-similarity inside an image patch. Consider that an image patch is divided into several smaller sub-regions. If these sub-regions are visually approximate to each other, then the image patch is of high consistency. Note that our proposed segmentation framework adopts a GAN-based augmentation following the patch selection stage. Considering the difficulty of image synthesis, higher intra-patch consistency that means less variations in texture can benefit the convergence of GANs to produce high-quality augmented samples. 

To search for the image patches with high representativeness and intra-patch consistency, we develop the Consistency-based Patch Selection (CPS) algorithm that is shown in Fig.~\ref{fig_cps} and Algorithm~\ref{algo1}. The algorithm consists of 3 stages: image patch sampling, dual-level clustering, and criterion computing.

\textbf{Image patch sampling} 
We crop a certain amount of $s\times s$ image patches from all images in the training set with a uniformly-sliding window of step $t$. These sampled image patches are fixed and no longer updated. The finally labeled image patches are only from these sampled ones and no more image patches will be sampled.

\textbf{Dual-level clustering} To estimate the representativeness of an image patch, we adopt K-Means algorithm~\cite{gao2020deep} to group image patches. To rate the intra-patch consistency, an image patch is split into sub-regions. We further group these sub-regions to obtain fine-grained representativeness. Thus, we conduct dual-level clustering that consists of a coarse level and a fine one. The feature vectors of image patches or sub-regions are computed by an ImageNet-pretrained~\cite{deng2009imagenet} ResNet50 model~\cite{he2016deep} which is able to extract generic representations~\cite{donahue2014decaf} for repetitive textures~\cite{cimpoi2014describing}. We adopt Euclidean distance of the extracted features for K-Means, inspired by the works~\cite{huang2014deep
,gao2020deep}. Using these features, we conduct the coarse-level clustering to divide the image patches into $K_1$ clusters. For each coarse-level cluster $C_k$, we cut each $s\times s$ image patch of $C_k$ into 2-by-2 sub-regions of size $\frac{s}{2}\times \frac{s}{2}$. Then we perform the fine-level clustering that groups the sub-regions of the same coarse-level cluster into $K_2$ clusters. In total, $K_1\!\times\! K_2$ fine-level clusters are obtained.

\begin{algorithm}[!tbp]
\renewcommand{\algorithmicrequire}{\textbf{Input:}}
\renewcommand{\algorithmicensure}{\textbf{Output:}}
\footnotesize
\caption{Consistency-based Patch Selection}
\begin{algorithmic}[1]
\Require {$\textit{Imgs}$ (a list of training images), $f$ (a feature extractor), $K_1$ (the number of coarse-level clusters), $K_2$ (the number of fine-level clusters in each coarse-level cluster). }
\Ensure {$\textit{Results}$ (a list of $K_1$ selected image patches).}

\Function {dual\_level\_clustering}{$Pats$,\, $K_1$,\, $K_2$,\, $f$}
    \State $c_{fine} = list()$
    \State $C_{coarse} = Kmeans(f(\textit{Pats}),\, n\_clusters=K_1)$
    \For {$k$  \textbf{in range} ($K_1$)}
        \State {$C_k = C_{coarse}[k]$}
        \State {$\textit{Regions} = crop2(C_k)$}
        \State {$c\_list = Kmeans(f(\textit{Regions}),\, n\_clusters=K_2)$}
        \State {$c_{fine}.append(c\_list)$}
    \EndFor
    \State \Return { $C_{coarse}$, $c_{fine}$}
\EndFunction
    
    \State $\textit{Pats} = crop1(\textit{Imgs})\;//\text{ Image patch sampling}$
    \State $C_{coarse}, c_{fine}=$ \Call{dual\_level\_clustering}{$\textit{Pats}$, $K_1$, $K_2$, $f$}
    \State $//\text{ } C_{coarse} \text{ is a list of } K_1 \text{ coarse-level clusters.}$
    \State $//\text{ } c_{fine} \text{ is a 2D list of } K_1\!\times\!K_2 \text{ fine-level clusters.}$
    \State $\textit{Results} = list()$
    \For {$k$  \textbf{in range} ($K_1$)}
        \State $C_k = C_{coarse}[k]$
        \State $\boldsymbol{v_{coarse}} = center\_vector\_of (C_k)$
        \State $c\_list = c_{fine}[k] \text{  }// \text{ a list of } K_2 \text{ fine-level clusters}$
        \State $c\_size\_list = list([element\_number(\boldsymbol{c'}) \textbf{ for } \boldsymbol{c'} \textbf{ in } c\_list])$
        \State {$\boldsymbol{c^*}=c\_list[argmax(c\_size\_list)]$}
        \State {$\boldsymbol{v_{fine}} = center\_vector\_of (\boldsymbol{c^*})$}
        \For {$\boldsymbol{x}$ \textbf{in} $C_k$}
            \State $C(\boldsymbol{x}) = \boldsymbol{v_{coarse}},\; c(\boldsymbol{x}) = \boldsymbol{v_{fine}}$
            \State $[\boldsymbol{x_1,x_2,x_3,x_4}] = crop2(\boldsymbol{x})$
            \State {$d_1 = \left\|f(\boldsymbol{x})-C(\boldsymbol{x})\right\|,\; d_2 = \frac{1}{4} \sum_{p=1}^{4} \| f\left(\boldsymbol{x_{p}}\right)-c(\boldsymbol{x}) \|$}
            \State {$d_3 = \max _{(i, j) \in
            \{\forall i, j: 1 \leq i<j \leq 4\}}
            \left\{\left\|f\left(\boldsymbol{x_{i}}\right)-f
            \left(\boldsymbol{x_{j}}\right)\right\|\right\} $}
            \State {$d.append(d_1 + d_2 + d_3)$}
        \EndFor
        \State {$\textit{Results}.append(\boldsymbol{x^k} = C_k[argmin(d)])$}
    \EndFor
    \State \Return $\textit{Results}$
\end{algorithmic}
\label{algo1}
\end{algorithm}

\textbf{Criterion computing} For each coarse-level cluster $C_k$, we traverse all $s\times s$ image patches in $C_k$ for one time, compute the selection criterion for each image patch, and search for the one that maximizes the coarse-to-fine representativeness as well as the intra-patch consistency. The selection criterion could be defined as Eq.~(\ref{CPS}):
\begin{align}
&\boldsymbol{x^k} = \mathop{\arg\min}_{\boldsymbol{x} \in C_k}\Big\{\|f(\boldsymbol{x})-C(\boldsymbol{x})\|+\frac{1}{4} \sum_{p=1}^4\left\|f\left(\boldsymbol{x_{p}}\right)-c(\boldsymbol{x})\right\| \notag \\
&+ \max _{(i,j)\in\{\forall{i,j}:1 \leq i<j \leq4\} } \left\{\left\|f\left(\boldsymbol{x_{i}}\right)-f\left(\boldsymbol{x_{j}}\right)\right\|\right\}\Big\}
\label{CPS}
\end{align}
where $\boldsymbol{x}$ denotes an $s\times s$ image patch. $\boldsymbol{x^k}$ is the selected image patch from $C_k$, the $k$-th coarse-level cluster. $f(\cdot)$ is the feature extractor that outputs a feature vector of the same size for different input sizes. $\boldsymbol{x_p/x_i/x_j}$ is one of the  $\frac{s}{2}\times\frac{s}{2}$ sub-regions within $\boldsymbol{x}$. $C(\boldsymbol{x})$ returns the center vector of the coarse-level cluster $C_k$ which $\boldsymbol{x}$ belong to. A \textit{center vector} is a vector that is the center of some cluster (a coarse-level or fine-level cluster). The center of a cluster is defined as the mean of all the elements (vectors) belonging to the cluster, following the K-means algorithm. From the $K_2$ fine-level clusters in $C_k$, we set $\boldsymbol{c^*}$ as the one with the largest number of sub-regions. $c(\boldsymbol{x})$ returns the center vector of $\boldsymbol{c^*}$. The selection criterion in Eq.~(\ref{CPS}) contains 3 cost terms that denote the coarse-level representativeness (the distance to the coarse-level cluster center), the fine-level representativeness (the summed distance from each sub-region to the largest fine-level cluster), and the intra-patch consistency (the maximum distance of any two sub-regions) respectively. Lower costs mean higher representativeness or consistency. After running the CPS algorithm, we choose an image patch for each coarse-level cluster, and obtain exactly $K_1$ image patches which are then annotated by experts.

\subsection{Conditional SinGAN Augmentation}\label{sec_csingan}
In this section, we propose a novel data augmentation strategy based on a Conditional Single-image GAN (CSinGAN). After the image patch selection in Section~\ref{sec_CPS}, we obtain $K_1$ pairs of image and mask. For each pair of image and mask, we propose a mask synthesis algorithm to produce lots of masks. Then the image patch, the real mask and these synthetic masks are employed to train the proposed CSinGAN. The trained generator synthesizes fake images corresponding to the synthetic masks. Following the above procedures, many pairs of synthetic image and mask are created from a real pair with our proposed CSinGAN.

\begin{figure}[!htb]
\includegraphics[width=1.0\linewidth]{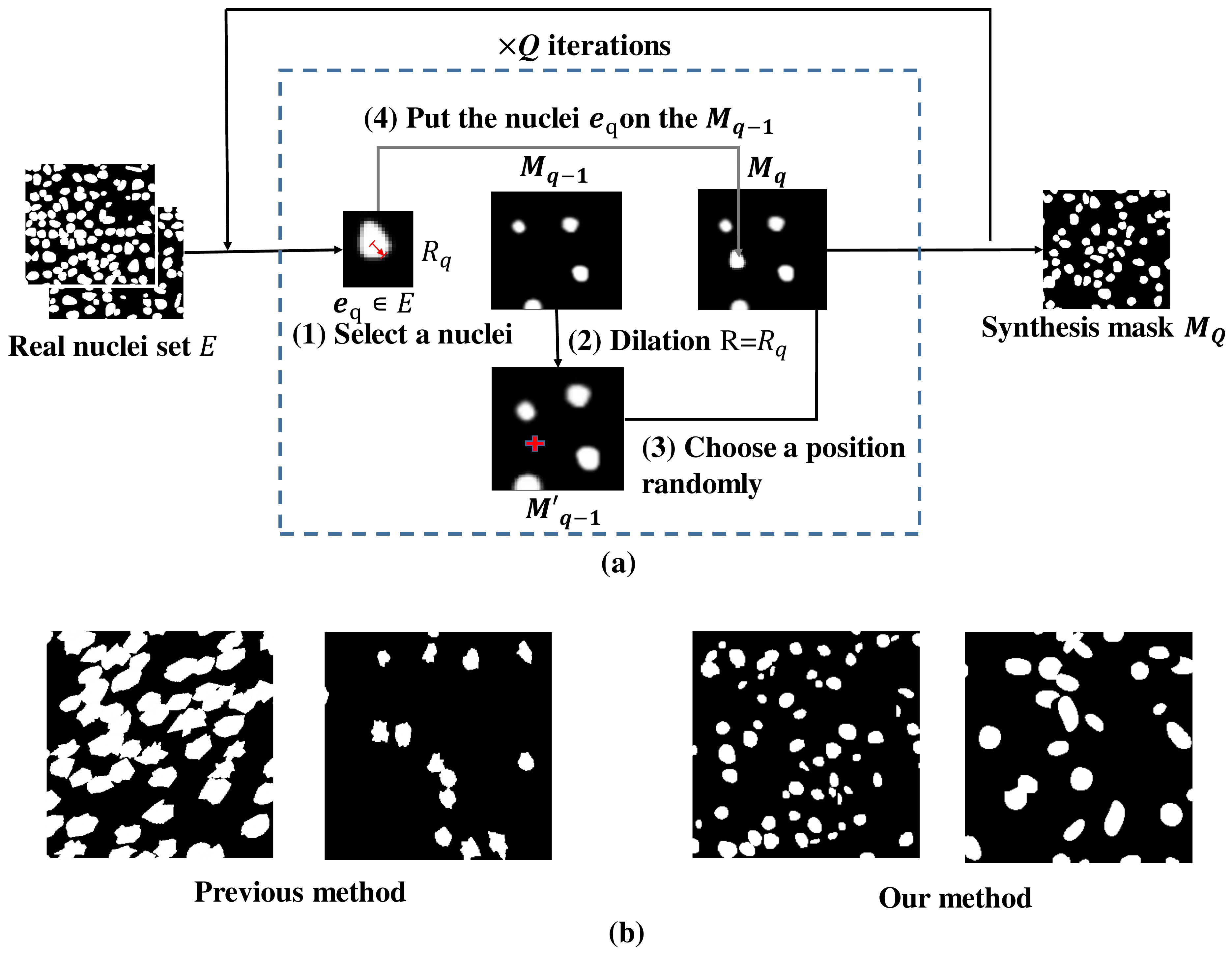}
\centering
\caption{Mask synthesis process and the synthetic results. (a) shows the synthesis process that has $Q$ iterations. Each iteration has four steps (1)-(4). (b) shows the synthesis results of an existing method \cite{Hou2019Robust} and our proposed mask synthesis.}
\label{fig_masksyn}
\end{figure}

To synthesize nuclei masks, Hou \textit{et al.} \cite{Hou2019Robust} sample random polygons from a pre-defined distribution which is built on statistical features such as radius, irregularity and spike of a polygon. However, the existing method requires a large number of mask annotations and it is difficult to estimate an accurate distribution of nucleus shape using a single mask. Therefore, we proposed a novel mask synthesis algorithm as shown in Fig.~\ref{fig_masksyn} (a).

Given one of the $K_1$ masks, we perform data augmentation (flip, random cropping, rotation) on the mask and collect all the augmented nuclei to form a set $E$. Then, we create a new empty $h' \times w'$ map. The main idea is iteratively choosing a nucleus from $E$ and placing the nucleus into the empty map, while ensuring that the placed nuclei are not overlapped with each other. The proposed mask synthesis algorithm has $Q$ iterations. At the $q$-th iteration, we first randomly select a nucleus denoted as $\boldsymbol{e_q}$ from $E$, and then calculate the maximum distance (denoted as $R_q$) from the nucleus center to the nucleus boundary. To avoid overlapping with the previously placed nuclei, we dilate $\boldsymbol{M_{q-1}}$ (the synthetic mask produced by the last iteration) with a radius of $R_q$. Next, we select a random position from the background (with black color in Fig.~\ref{fig_masksyn}) of the dilated mask, and insert $\boldsymbol{e_q}$ into $\boldsymbol{M_{q-1}}$ at the selected position. After $Q$ iterations, we obtain a mask map $\boldsymbol{M_{Q}}$ with $Q$ nuclei, and then crop the map to a smaller size $h \times w$ to simulate the incomplete nuclei in real masks. As shown in Fig.~\ref{fig_masksyn} (b), with the same limited number of nuclei annotations our proposed mask synthesis algorithm generates more realistic shapes, while the existing method~\cite{Hou2019Robust} produces nuclei of unsmoothed boundaries.

\begin{figure*}[!htb]
\includegraphics[width=0.9\linewidth]{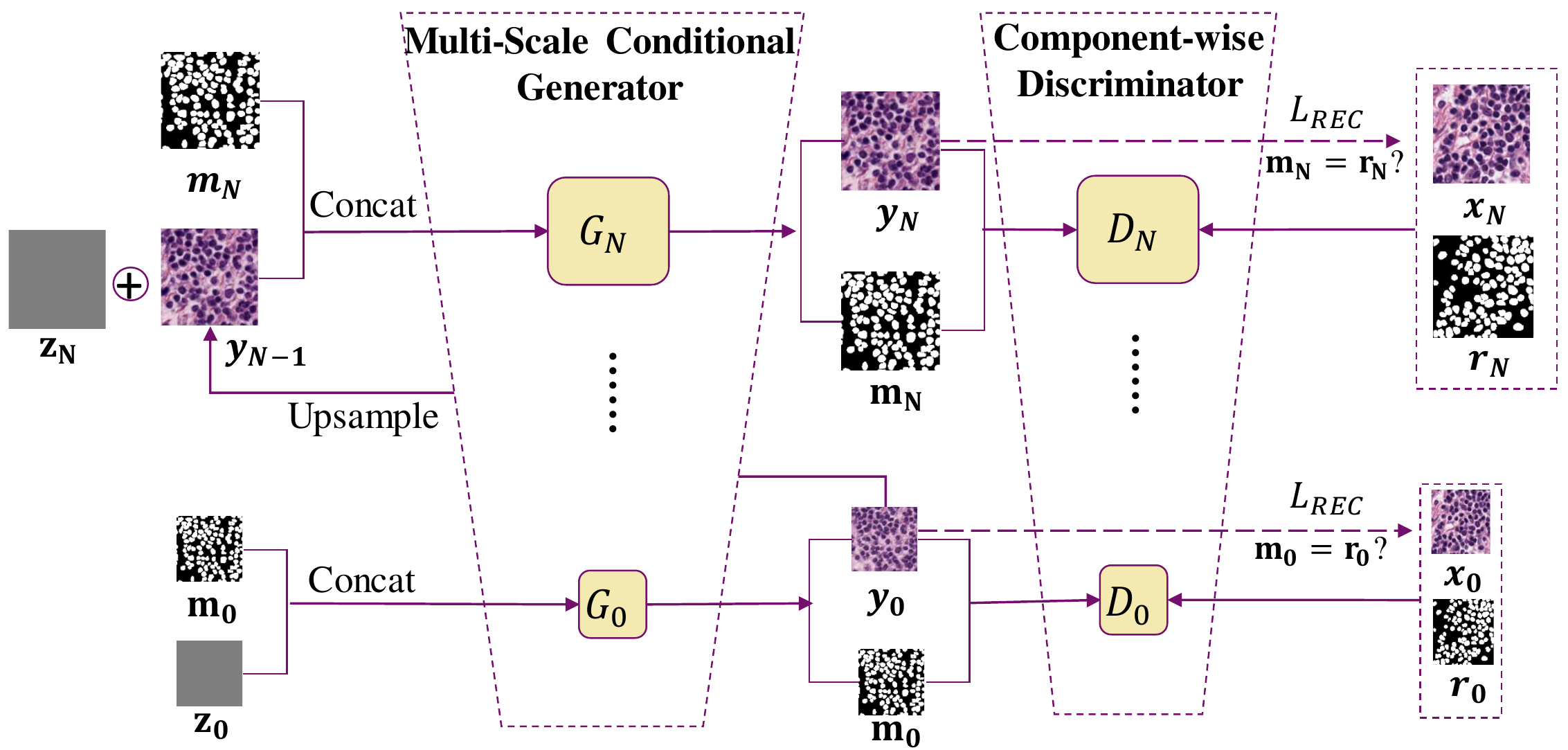}
\centering
\caption{Conditional Single-image GAN (CSinGAN). CSinGAN contains a multi-scale conditional generator and a component-wise discriminator. $\boldsymbol{m_N}$ is sampled from real and synthetic labels. The reconstruction loss $L_{REC}$ is computed when $\boldsymbol{m_N}$ is real. }
\label{fig_csingan}
\end{figure*}

To generate corresponding nuclei images for the synthetic masks, we introduce a novel conditional SinGAN that consists of a multi-scale conditional generator and a novel component-wise discriminator, as shown in Fig.~\ref{fig_csingan}. The generator and the discriminator are denoted as $\left\{G_{0},...,G_{N}\right\}$ and $\left\{D_{0},...,D_{N}\right\}$ respectively. The multi-scale generator could be defined recursively as:
\begin{equation}
\boldsymbol{y_n}=
\begin{cases}
G_n([\boldsymbol{m_n, z_n+y_{n-1}}]),\quad 0<n\leq N \\
G_0([\boldsymbol{m_0, z_0}]),\qquad\qquad\qquad\; n=0
\end{cases}
\end{equation}
where $\boldsymbol{m_n}$ and $\boldsymbol{m_0}$ denote a synthetic mask or the real one. When computing $\boldsymbol{y_n}$, all $\boldsymbol{m_i} (0\leq i\leq n)$ are obtained by resizing the same mask $\boldsymbol{m}$ to different scales. $\boldsymbol{m}$ is randomly sampled from the real mask and the synthetic masks. $\boldsymbol{z_n}$ and $\boldsymbol{z_0}$ denote a 3-channel Gaussian noise and $\boldsymbol{z_n}$ has the same shape as $\boldsymbol{y_{n-1}}$. $[\cdot,\cdot]$ means concatenating two variables along the channel dimension. The internal architecture of $G_n$ is based on~\cite{Shaham2019Singan}. At each scale the training loss contains an adversarial term and a reconstruction term: $\min _{G_{n}} \max _{D_{n}} \mathcal{L}_{\mathrm{ADV}}\left(\boldsymbol{x_{n},y_{n},m_{n},r_{n}}, D_{n}\right)+\alpha \mathcal{L}_{\mathrm{REC}}\left(G_{n}\right)$. The reconstruction loss is computed as $\mathcal{L}_{\mathrm{REC}}=\|\boldsymbol{y_n-x_n}\|^2$ by setting $\boldsymbol{m_n}$ as the real mask, $\boldsymbol{z_n}$ as zeros, $\boldsymbol{x_n}$ as the real image.

For the adversarial loss, we develop a novel component-wise discriminator (CwD) that separates an image into different components and classifies them respectively. The discriminator $D_n$ has a foreground sub-net $D_f$, a background sub-net $D_b$ and a global one $D_g$. By setting all $\boldsymbol{m_n}$ as synthetic masks, the overall adversarial loss is calculated as:
\begin{align}
\mathcal{L}_{\mathrm{ADV}}=
&\beta \mathcal{L}_{\mathrm{adv}}\left( D_{g}, \boldsymbol{y_n, x_n} \right) + \notag \\
&\gamma \mathcal{L}_{\mathrm{adv}}\left( D_{f}, \boldsymbol{y_n} \otimes \boldsymbol{m_n, x_n} \otimes \boldsymbol{r_n} \right) +
\notag \\
&\delta \mathcal{L}_{\mathrm{adv}}\left( D_{b}, \boldsymbol{y_n} \otimes \neg \boldsymbol{m_n, x_n} \otimes \neg \boldsymbol{r_n} \right)
\end{align}
where $\boldsymbol{r_n}$ denotes the real mask resized to scale $n$. $\otimes$ denotes element-wise multiplication and can be viewed as a masking operation. $\boldsymbol{x_n}\otimes \boldsymbol{r_n}$ and $\boldsymbol{x_n} \otimes \neg \boldsymbol{r_n}$ mean extracting the foreground region and the background respectively for $\boldsymbol{x_n}$. Thus different sub-nets could focus on different components. $D_g$, $D_f$ and $D_b$ do not share weights but adopt the same architecture following~\cite{Shaham2019Singan}.

$\mathcal{L}_{\mathrm{adv}}$ denotes WGAN-GP loss~\cite{Gulrajani2017Improved}. WGAN-GP loss is a loss for generative adversarial networks that augments the Wasserstein GAN (WGAN) loss~\cite{Arjovsky2017Wasserstein} by imposing a gradient norm penalty on random samples to achieve Lipschitz continuity. The training of the WGAN-GP loss requires very little hyper-parameters tuning, and is more stable than that of the WGAN loss. The detailed WGAN-GP loss is shown as Equation~(\ref{eq_wgangp}):
\begin{align}
\mathcal{L}_{\mathrm{adv}} \left( D, \boldsymbol{y}, \boldsymbol{x} \right) = &{\mathbb{E}}[D({\boldsymbol{y}})]-{\mathbb{E}}[D(\boldsymbol{x})]+ \notag \\
\lambda &{\mathbb{E}}\left[\left(\left\|\nabla_{\hat{\boldsymbol{x}}} D(\hat{\boldsymbol{x}})\right\|_{2}-1\right)^{2}\right]
\label{eq_wgangp}
\end{align}
where $\boldsymbol{y}$ and $\boldsymbol{x}$ mean a generated image and the real image. $\hat{\boldsymbol{x}}$ is uniformly sampled along straight lines between two points which are sampled from the distribution of $\boldsymbol{x}$ and $\boldsymbol{y}$ respectively. $D(\cdot)$ is a discriminator. $\mathbb{E}\left[\cdot \right]$ denotes the expectation. $\lambda$ is the penalty coefficient to weight the gradient penalty term. 

\subsection{Semi-supervised Nuclei Segmentation}\label{sec:semi}
In this section, we describe how to train a CNN model for nuclei segmentation. After the augmentation in Section~\ref{sec_csingan}, we have a few real pairs of image and mask, many synthetic pairs and lots of unlabeled images. The fake pairs can be regarded as labeled images so the task is formulated as a semi-supervised problem. In this paper semi-supervised learning (SSL) is not a contribution but a training paradigm of our proposed framework, and we simply adopt a classical SSL method pseudo-labeling. Pseudo-labeling methods~\cite{lee2013pseudo} usually apply a trained model to predict results for unlabeled data. The predicted results can be viewed as and converted into the `ground-truth' labels of the unlabeled data. These predicted `ground-truth' annotations are referred to as pseudo labels which can be used to re-train or fine-tune the model.

To segment nuclei, the proposed framework employs a popular instance segmentation model, Mask-RCNN~\cite{He2017Mask}. First, the prediction model (e.g. Mask-RCNN) is trained with the real and fake pairs. Then the model predicts nuclei masks for each image of the trainset. These predicted masks are saved as the `ground-truth' annotations, namely, the pseudo labels. The pairs of pseudo labels and their corresponding images are added to the training set for the next round of prediction model training. The performance is not improved any more after 2-3 iterations. The last model is chosen as the final model. We claim that the proposed label-efficient framework could also work with other existing nuclei segmentation models based on CNN, like Hover-net. Other SSL methods that are proposed for instance segmentation could be directly applied to our proposed segmentation framework. 

\begin{table*}[!t]
\centering
\setlength{\tabcolsep}{5mm}{
\begin{tabular}{c|c|c|l} \hline
Methods & AJI & Dice & \quad  $P$-value\\ \hline
RndCrop + MRCNN & 0.4376 & 0.6374 & $9.76 \times 10^{-23}$\\ 
RndCenCrop + MRCNN &  0.4637&  0.6860 & $4.82 \times 10^{-18}$\\ \hline
K-means + MRCNN &  0.4724 &  0.7093  & $2.17 \times 10^{-4}$\\ 
CPS w/o 3rd term + MRCNN &  0.4825 &  0.6962 &  $9.03 \times 10^{-8}$\\ 
CPS w/o 2nd term + MRCNN &  0.4735 &  0.6863 &  $7.63 \times 10^{-6}$ \\ 
CPS + MRCNN &  \textbf{0.4920} &  \textbf{0.7105} & \qquad  - \\ \hline
K-means + MRCNN + Plabel & 0.5097 & 0.7328 & $5.60 \times 10^{-29}$\\
CPS w/o 3rd term + MRCNN + Plabel & 0.5289 & 0.7403 & $3.43 \times 10^{-9}$ \\
CPS w/o 2nd term + MRCNN + Plabel & 0.5264 & 0.7373 & $4.82 \times 10^{-12}$ \\
CPS + MRCNN + Plabel &\textbf{0.5374} & \textbf{0.7536} & \qquad - \\\hline
\end{tabular}}
\caption{Effectiveness of the proposed Consistency-based Patch Selection (CPS) algorithm. Each patch selection method is followed by the proposed CSinGAN augmentation and then a Mask-RCNN is trained and tested to obtain the above results. RndCrop denotes the mean performance of four random seeds (21,100,500,1000). MRCNN and Plabel donate Mask-RCNN\cite{He2017Mask} and Pseudo-labeling respectively.}
\label{tab1_CPS}
\end{table*}

\section{Experiments}
\subsection{Implementation Details}
\textbf{Datasets}
We conduct experiments on three datasets: TCGA-KUMAR database~\cite{Kumar2017adataset}, TNBC~\cite{Naylor2018Segmentation} and MoNuSeg~\cite{kumar2019multi}. The TCGA-KUMAR dataset has 30 labeled images of size $1000 \times 1000$ at 40$\times$ magnification. It is obtained from The Cancer Genome Atlas (TCGA) and each image is from one of the seven organs, including breast, bladder, colon, kidney, liver, prostate, and stomach. We split the data into 12 training images, 4 validation images and 14 testing images. following~\cite{liu2021panoptic,Naylor2018Segmentation}. The TNBC dataset, which contains 50 annotated images of size $500 \times 500$, is collected from 11 different patients of the Curie Institute. The dataset was split into 30 training images, 7 validation images, and 13 test images in our experiments. The MoNuSeg dataset consists of 44 labeled images, 30 for training and 14 for testing. The image size of MoNuSeg is $1000 \times 1000$.

\textbf{Preprocessing}
Following the previous work, all images were cropped into $256 \times 256$ patches for training the network efficiently. Naive data augmentation strategies including rotation (\ang{90},\ang{180},\ang{270}), flip, gaussian blur, gaussian noise are applied to these cropped image patches.

\textbf{Patch Selection}
For consistency-based patch selection (CPS), we set $s=256$, $t=15$, $K_1=6/9/18$ (corresponding to 5\%/5\%/7\% labels of the trainset) for the TNBC/TCGA-KUMAR/MoNuSeg dataset respectively, $K_2=4$. The training set of these datasets are cropped into $256 \times 256$ image patches with stride 15. The CPS algorithm takes around 1-3 hours for each dataset.

\textbf{Paired Sample Synthesis}
In each scale of the proposed CSinGAN, the generator consists of 5 convolution blocks. Each block contains a convolution layer of kernel size $3 \times 3$, a BatchNorm layer and a LeakyReLU function. The channel number of each block starts from 32 and doubles every 4 scales. The multi-scale discriminators have the same architecture as the generators but they are trained with a different loss from the generators. We use $K_1$ CsinGANs and each CsinGAN augments data from a representative image patches. For each CSinGAN, 50 synthetic masks of size $256 \times 256$ are synthesized for training the nuclei segmentation model. For the adversarial loss of CSinGAN, we set $\beta, \gamma, \delta, \lambda$ to 1.0, 1.0, 1.0, 0.1. We follow a related work SinGAN~\cite{Shaham2019Singan} to set the above hyper-parameters.

\textbf{Semi-supervised Nuclei Segmentation Baselines}
Mask-RCNNs are trained for 200 epochs with an SGD optimizer with an NVIDIA V-100 GPU, an initial learning rate of 0.005, a weight decay of $10^{-4}$, a momentum of 0.9, a batch size of 6, and the input size $256 \times 256$. During the testing stage the input size is set as the original image size of each dataset. Hover-nets~\cite{Graham2019Hover} are trained for 100 epochs with an Adam optimizer with a V-100 GPU, an initial learning rate of $10^{-4}$. At the begining of the training, all models are initialized with the model weights pre-trained on ImageNet~\cite{deng2009imagenet}. For each round of semi-supervised learning, 600 pseudo labels are predicted by the last supervised segmentation model and added to the training data. The time cost of the semi-supervised training is 1-2 days.

\begin{figure*}[!h]
\includegraphics[width=1.0\linewidth]{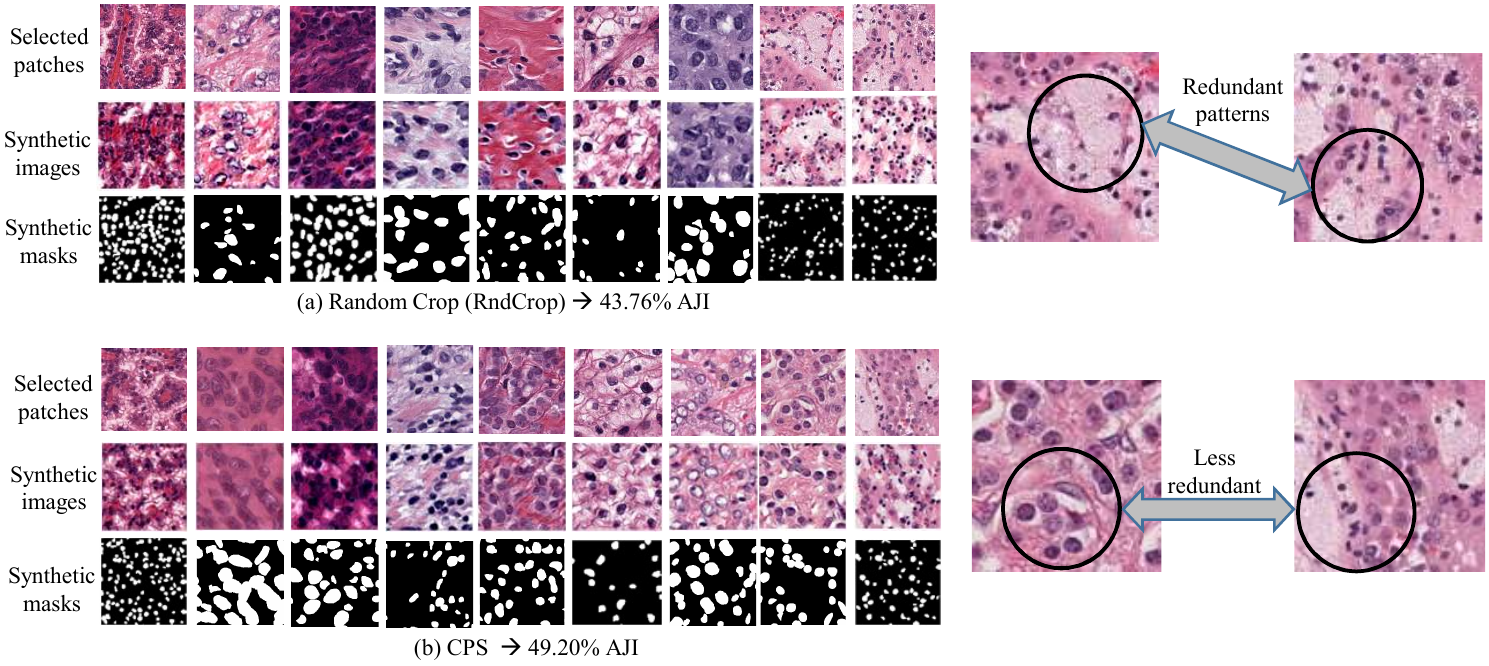}
\centering
\caption{Visualized comparison between the proposed consistency-based patch selection and the random crop method. (a) shows that the random crop method may choose two image patch which have redundant patterns. (b) displays that the image patches selected by the proposed CPS contain less redundant textures, and can better replace the original trainset.}
\label{fig_vis}
\end{figure*}

\subsection{Evaluation Criteria}
To evaluate nuclei segmentation models, both pixel-level and object-level criteria are applied. The object-level evaluation is for cell nucleus localization, and the pixel-level correlation evaluation is for preserving fine boundaries. In our quantitive study, we use two different metrics, Aggregated Jaccard Index (AJI)~\cite{Kumar2017adataset} and Dice Coefficient.

\textbf{Aggregated Jaccard Index (AJI)}: AJI~\cite{Mahmood2019Deep} is an extension of the global Jaccard index. It is defined as: $AJI=\frac{\sum_{i=1}^{ N\!P }\left|\boldsymbol{T_{i}} \cap \boldsymbol{S_{j}^{*}(i)}\right|}{\sum_{i=1}^{N\!G}\left|\boldsymbol{T_{i}} \cup \boldsymbol{S_{j}^{*}(i)}\right|+\sum_{g \in U}\left|\boldsymbol{S_{g}}\right|}$ where $\boldsymbol{S_{j}}$ is a prediction result and $\boldsymbol{T_{i}}$ is the ground truth.  $\boldsymbol{S_{j}^{*}(i)}$ represents the connected component that has the maximum intersection with the ground truth. $U$ is the set of areas that do not overlap with any real nuclei. $N\!G$, $N\!P$ are the number of ground truths and predictions respectively. AJI measures the overlapping areas of multiple objects and is recognized as an object-level criterion for segmentation evaluation. 

\textbf{Dice Coefficient (Dice)}: Dice is a typical pixel-level metric for validating medical image segmentation, and is computed as $\frac{2|\boldsymbol{X} \cap \boldsymbol{Y}|}{|\boldsymbol{X}|+|\boldsymbol{Y}|}$. $\boldsymbol{X}$, $\boldsymbol{Y}$ denote ground truths and predictions respectively. A higher Dice value indicates that the predicted segmentation mask has a larger intersection with the ground truth, and means that the segmentation result is more accurate.

\subsection{Effectiveness of Consistency-based Patch Selection}
This section shows if our proposed consistency-based patch selection (CPS) is effective. TABLE~\ref{tab1_CPS} presents the performance of different patch selection methods on the TGCA-KUMAR dataset. The samples chosen by each selection method are augmented by the proposed CSinGAN. Then the augmented samples are used to train a Mask-RCNN whose testing results are logged as the performance of the selection method. In TABLE~\ref{tab1_CPS}. our proposed CPS is compared with two kinds of random selections, RndCrop and RndCenCrop. The RndCrop strategy cuts out many $s\times s$ image patches from the whole trainset, and then randomly chooses $K_1$ patches to annotate. The RndCenCrop strategy randomly selects $K_1$ full images from the trainset, and crops the centering $s\times s$ image patch for each selected full image. The random seeds (Seed1-Seed4) used in these experiments are 21, 100, 500, 1000. TABLE~\ref{tab1_CPS} shows the CPS significantly outperforms these random selections by about 2\% - 6\% AJI. 

\begin{table}[!ht]
\centering
\setlength{\tabcolsep}{3mm}{
\begin{tabular}{c|c|c|c}
\hline
Methods & AJI & Dice&$P$-value \\ \hline
NaïveAug & 0.4786 & 0.6898 &  $7.24 \times 10^{-17}$\\ 
pix2pix \cite{Isola2017Image} &  0.4386 & 0.6689 & $3.39 \times 10^{-28}$ \\ 
cycleGAN \cite{Zhu2017Unpaired} &  0.4828& 0.7014 & $4.49 \times 10^{-11}$\\ 
SinGAN \cite{Shaham2019Singan} &0.3968 & 0.6064 & $2.38 \times 10^{-28}$ \\ 
CSinGAN(w/o CwD) & 0.4730 & 0.6882& $4.71 \times 10^{-8}$ \\ 
CSinGAN & \textbf{0.4920} & \textbf{0.7105}& - \\ \hline
\end{tabular}}
\caption{Effectiveness of the proposed Conditional SinGAN augmentation on the TCGA-KUMAR dataset. Each augmentation strategy is based on the samples selected by our proposed CPS algorithm. These augmentation methods are evaluated by training a Mask-RCNN on their synthesized samples and testing the Mask-RCNN. CwD denotes the proposed component-wise discriminator. The CSinGAN (w/o CwD) does not use our proposed component-wise discriminator, but only uses a single discriminator with the standard adversarial loss.}
\label{tab2_CSinGAN}
\end{table}
	
\begin{table*}[!ht]
\centering
\setlength{\tabcolsep}{5mm}{
\begin{tabular}{c|c|c|c|c}
\hline
Methods & AJI-s & AJI-u & AJI & Dice  \\ \hline
a).RndCrop + MRCNN & 0.4371 & 0.4395 & 0.4381  & 0.6499 \\ 
b).RndCenCrop + MRCNN & 0.4700 & 0.4321 & 0.4537  & 0.6662 \\ 
c).RndCrop + NaïveAug + MRCNN & 0.4673 & 0.4481 & 0.4577  & 0.6638 \\ 
d).RndCenCrop + NaïveAug + MRCNN & 0.4701 &0.4623 & 0.4662  & 0.6808\\ 
e).RndCrop + NaïveAug + MRCNN + Plabel & 0.5182 & 0.5160 & 0.5171  & 0.7282\\ 
f).RndCenCrop + NaïveAug + MRCNN + Plabel & 0.5236 & 0.5134 & 0.5185  & 0.7255 \\ 
\textbf{Ours} &0.5348 &\textbf{0.5408} & \textbf{0.5374} & \textbf{0.7536} \\ \hline
Fully-supervised Mask-RCNN~\cite{He2017Mask} &\textbf{0.5389} &0.5332 & 0.5364\footnotemark[1]&0.7459\\ \hline
\end{tabular}}
\caption{Effectiveness of our proposed label-efficient nuclei segmentation framework (denoted as `Ours') on the TCGA-KUMAR dataset. AJI-s/AJI-u denote AJI on seen/unseen categories (organs). MRCNN and Plabel denote Mask-RCNN\cite{He2017Mask} and Pseudo-labeling respectively.}
\label{tab_full}
\end{table*}
	
\begin{table*}[!t]
\centering
\setlength{\tabcolsep}{3mm}{
\begin{tabular}{c|c|c|c|c|c|c|c|c|c}
\hline
&\multicolumn{3}{|c|}{TCGA-KUMAR} &\multicolumn{3}{|c}{TNBC} &\multicolumn{3}{|c}{ MoNuSeg }\\ \hline
Methods & AJI & Dice & $P$-value & AJI & Dice & $P$-value & AJI & Dice & $P$-value \\ \hline
MRCNN~\cite{He2017Mask} & 0.5364 & 0.7459 & $6.75\times 10^{-18}$ & 0.5279 & 0.6978 & $2.84\times 10^{-27}$ & 0.5950 & 0.7677 & $4.79\times 10^{-20}$ \\
DIST \cite{Naylor2018Segmentation} & 0.5598 & 0.7863 & $8.68\times 10^{-19}$ & 0.5258 & 0.7368 & $3.43\times 10^{-20}$ & 0.5839 & 0.7589 & $7.83\times 10^{-18}$\\
Micro-Net \cite{raza2019micro} & 0.5631 & 0.7961 & $9.61\times 10^{-16}$ & 0.5064 & 0.6848 & $1.39\times 10^{-18}$ & 0.6011 & 0.7762 & $2.18\times 10^{-21}$ \\
MDC\_NET \cite{liu2021mdc} & 0.5803 & - & - & 0.6103 & - & - & - & - & - \\
DPMSFF \cite{Liu2019Nuclei} & 0.5854 & 0.7936 & - & 0.5865 & 0.7792 & - & - & - & - \\
PFFNET \cite{liu2021panoptic} & 0.5980\footnote[2]{} & 0.7981\footnote[2]{} & $6.86\times 10^{-23}$ & 0.6283\footnote[2]{} & \textbf{0.8127\footnote[2]{}} & $3.38 \times 10^{-20}$ & 0.6209 & 0.7878 & $3.77\times 10^{-7}$\\
Hover-net \cite{Graham2019Hover}
& \textbf{0.6107} & \textbf{0.8211} & $6.87\times 10^{-4}$ & \underline{0.6307} & 0.7819 & $4.36\times 10^{-8}$ & \textbf{0.6602} & \textbf{0.8213} & $5.73\times 10^{-35}$\\
\hline
\textbf{MRCNN + Ours} & 0.5374 & 0.7536 & $4.78\times 10^{-23}$ & 0.5281 & 0.6994 & $7.81\times 10^{-35}$ & 0.5906 & 0.7626 & $4.71\times 10^{-15}$ \\
\textbf{Hover-net + Ours} &  \underline{0.6086} & \underline{0.8188} & - & \textbf{0.6414} & \underline{0.7938} & - & \underline{0.6485} & \underline{0.8129} & - \\ \hline
\end{tabular}}
\caption{Effectiveness of the proposed framework on TCGA-KUMAR, TNBC and MoNuSeg datasets. MRCNN+Ours and HoverNet+ours mean a Mask-rcnn and a Hover-net with our proposed framework respectively. The methods denoted as `* + Ours' only adopt about 5\%/5\%/7\% pixel-level annotations of TCGA-KUMAR/TBNC/MoNuSeg datasets respectively. Other state-of-the-art models are supervised with all the training labels. Smaller $P$-value of a model means that the performance difference between the model and `Hover-net+ours' is more significant. The best result is in bold while the second best is underlined.}
\label{tab4_generalization}
\end{table*}

We conduct ablation studies for the three-term selection function (Eq.~(\ref{CPS})) in CPS by removing some terms to study their effectiveness. K-means denotes the method removing the 2nd term and the 3rd term in Eq.~(\ref{CPS}). We also test only removing the 2nd or the 3rd term. As TABLE~\ref{tab1_CPS} shows, without using the pseudo-label method, CPS outperforms Kmeans by 1.96\% AJI, CPS w/o 2nd/3rd term by 1.85\%/0.95\% AJI respectively. With the pseudo-labeling method, the proposed CPS still surpasses the other three methods by 2.77\%, 1.10\% and 0.85\% AJI respectively. These results suggest that both fine-grain representativeness (the 2nd term) and intra-patch consistency (the 3rd term) are beneficial for sample selection. Besides, we perform statistical significance analysis of AJI improvements by using paired t-test on TCGA-KUMAR dataset. The $P$-values in TABLE~\ref{tab1_CPS} show if the AJI differences between our proposed CPS method and other approaches is of statistical significance. No matter whether using the pseudo-labeling method, all the performance improvements are statistically significant with a $P$-value $<$ 0.05.

Fig.~\ref{fig_vis} illustrates that our CPS method chooses a training subset with less redundancy. Fig.~\ref{fig_vis}(a) visualizes the image patches selected by the RndCrop strategy. Fig.~\ref{fig_vis}(b) shows the CPS-selected image patches. In Fig.~\ref{fig_vis}(a), the 8th and the 9th image patches selected by the RndCrop method have very similar patterns. In contrast, the patches chosen by CPS show higher diversity than those selected by RndCrop, which may better represent the original trainset.

\subsection{Effectiveness of Conditional SinGAN Augmentation}
In this section the proposed CSinGAN is compared to other data augmentations on the TCGA-KUMAR dataset, shown in TABLE~\ref{tab2_CSinGAN}. NaiveAug is a set of transformations (rotation, flipping, color jittering, ...). Pix2pix~\cite{Isola2017Image} learns to translate $K_1$ masks to $K_1$ image patches. By taking real and synthetic masks as a domain, cycleGAN~\cite{Zhu2017Unpaired} attempts to map the domain to a domain of nuclei image. The `paint2image' application of SinGAN~\cite{Shaham2019Singan} could convert colored masks to images, which is denoted as `SinGAN' in TABLE~\ref{tab2_CSinGAN}. With CPS-selected image patches, the above methods synthesize samples to train a Mask-RCNN whose testing results reflect the strength of the corresponding augmentation strategy. TABLE~\ref{tab2_CSinGAN} displays that our proposed CSinGAN surpasses the existing methods by 1\% to 9.6\% AJI. 
\footnotetext[1]{AJI of other MRCNN reproductions: 0.5396~\cite{liu2021panoptic} / 0.5382~\cite{Liu2019Nuclei} / 0.5002~\cite{Naylor2018Segmentation}}

\footnotetext[2]{Published AJI/Dice performance on TCGA-KUMAR dataset: 0.6107/0.8091, on TNBC dataset: 0.6313/0.8037 of PFFNET\cite{liu2021panoptic}}
\begin{figure*}[h]
	\includegraphics[width=1.\linewidth, trim=0 0 150 0, clip]{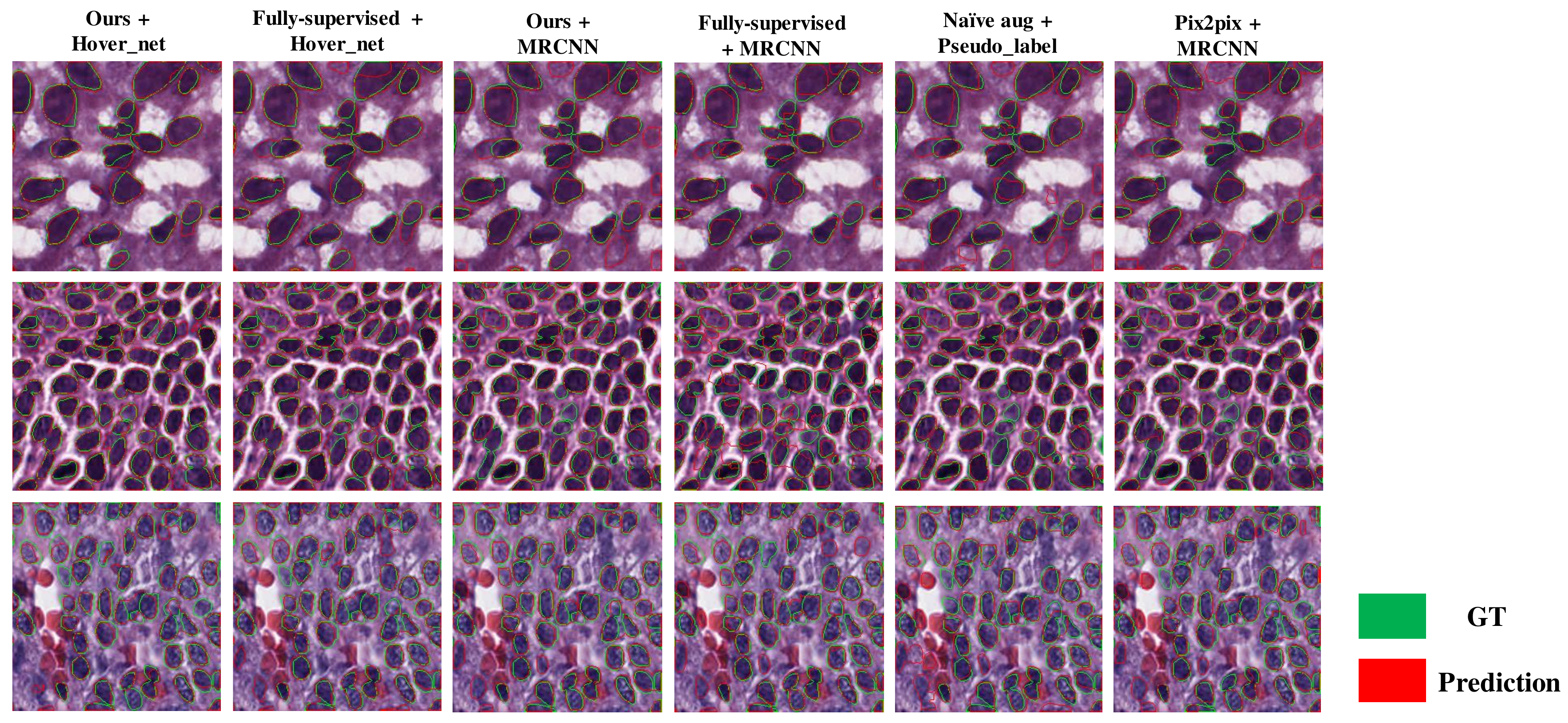}
	\centering
	\caption{Visualized comparisons between our proposed method and other existing models. The ground truth nuclei are marked with green boundaries and the predicted ones are marked with red color.}
	\label{fig_segvis}
\end{figure*}

\begin{table*}[!htb]
	\centering
	\setlength{\tabcolsep}{5mm}{
	\begin{tabular}{c|c|c|c|c|c|c|c}
		\hline
		Organs & Stomach & Prostate & Bladder & Liver & Kidney & Breast& Colon \\ \hline
		MRCNN+Ours &  0.6019 & 0.5561 & 0.5549 & 0.4902 & 0.5806 &  0.5123 &0.4657\\ 
		MRCNN \cite{He2017Mask} &0.6043 & 0.5850 & 0.5599& 0.4685 & 0.5677 & 0.5333 & 0.4355 \\ 
		Hover-net+Ours &  0.7091 & 0.6204 & 0.6017 & 0.5712 & 0.6415 & 0.6006 &0.5159\\ 
		Hover-net \cite{Graham2019Hover} &0.7158 & 0.6435 & 0.6113 & 0.5726 & 0.6312 & 0.6140 & 0.4924 \\ \hline
	\end{tabular}}
	\caption{Organ-wise AJI results on the TCGA-KUMAR dataset. MRCNN+Ours and Hover-net+Ours represent a Mask-rcnn and a Hover-net with our proposed framework respectively.}
	\label{tab_organ}
\end{table*}

\subsection{Effectiveness and Generalization of the Proposed Framework}
TABLE~\ref{tab_full} verifies the effectiveness of our proposed method on the TCGA-KUMAR dataset. `Ours' denotes the full proposed framework trained with a Mask-RCNN as the segmentation model, and no more than 5\% pixel-wise labels. The actual labeling ratio is $(256^2\times 9)/(1000^2\times 12)\approx 4.93\%$. `Plabel' denotes pseudo-labeling~\cite{lee2013pseudo} that is a simple semi-supervised method adopted by our proposed framework and is well described in Section~\ref{sec:semi}. As TABLE~\ref{tab_full} displays, our proposed method obtains even slightly better AJI and Dice scores than a full-supervised Mask-RCNN, by only annotating 5\% pixels. Interestingly, the fully-supervised MRCNN achieves higher AJI on seen categories while our proposed framework performs better with unseen categories. It indicates that the fully-supervised model may over-fit to some degree and that our framework has better generalization and higher robustness. In TABLE~\ref{tab_full} the proposed framework also outperforms the combinations of some existing methods which are denoted as a) - f).

To understand the generalization of our proposed framework, we further evaluate the framework with another state-of-the-art nuclei segmentation model, Hover-net~\cite{Graham2019Hover}, on TCGA-KUMAR, TNBC and MoNuSeg datasets. As Table~\ref{tab4_generalization} shows, our proposed framework using Hover-net and 5\% labels achieves almost the same AJI value with a fully-supervised Hover-net on the TCGA-KUMAR dataset. On the TNBC dataset, our proposed method with Hover-net is trained with only 5\% annotations, and even outperforms the fully-supervised Hover-net by 1\% AJI. On the MoNuSeg dataset, the proposed framework using MRCNN and 7\% labels is only 0.5\% AJI \& Dice lower than the fully-supervised MRCNN. These results indicate that our method could effectively work with different nuclei segmentation models and datasets. Besides, Hover-net+ours also surpasses some state-of-the-art nuclei segmentation models, PFFNET~\cite{Liu2019Nuclei} and MDC-NET~\cite{liu2021mdc}, by 1\% and 3\% AJI on the TNBC dataset. The $P$-values in TABLE~\ref{tab4_generalization} show if the AJI differences between Hover-net+Ours and other methods are statistically significant on these datasets. 

To understand the multi-organ performance of our proposed framework, we report the organ-wise results on the TCGA-KUMAR dataset. As shown in TABLE~\ref{tab_organ}, our proposed approach does not favor some specific organs or cancer-types. For example, for liver MRCNN+Ours outperforms MRCNN by 2.17\% AJI while Hover-net+Ours is worse than Hover-net. However, we find that when using our proposed approach, the variation of AJI results among different organs is relatively smaller. For example, the AJI variation of different organs is 13.6\% (46.6\%-60.2\%) for the MRCNN using our proposed approach. For the MRCNN without using our method, the AJI variation is 16.8\% (43.6\%-60.4\%). Similarly, the AJI variation of different organs is 19.3\% (51.6\%-70.9\%) for the Hover-net+Ours. For the Hover-net without our approach, the AJI variation is 22.4\% (49.2\%-71.6\%). The above results could indicate that using our proposed method leads to smaller variation and better generalization across different organs.

Fig.~\ref{fig_segvis} presents some of the visualized segmentation results on the TCGA-KUMAR dataset. The ground truth areas are surrounded by green boundaries while the predicted nuclei are marked with red color. Ours+Hover\_net and Ours+MRCNN denote our proposed framework with these two segmentation models using less than 5\% annotations. Fully-supervised Hover\_net and MRCNN represent the segmentation models using 100\% training annotations. NaiveAug+PseudoLabel denotes a Mask-RCNN trained with naive data augmentation and pseudo labeling while Pix2pix+MRCNN denotes the Mask-RCNN that uses a pix2pix model to augment data. As Fig.~\ref{fig_segvis} displays, NaiveAug+PseudoLabel and Pix2pix+MRCNN predict loose nucleus regions, while our methods with HoverNet and MRCNN locate nuclei more accurately.

\subsection{Investigation of Hyper-parameters}
\begin{table}[!t]
\centering
\setlength{\tabcolsep}{5mm}{
\begin{tabular}{p{0.4cm}<{\centering}|c|c|c} \hline
Annos & $K_2$ = 4 & AJI & Dice\\ \hline
\multirow{2}*{$\sim$3\%}
& $K_1$ = 6 w/o CSinGAN & 0.4317 & 0.6428\\
~ & $K_1$ = 6 w/ CSinGAN & \textbf{0.4517} & \textbf{0.6728}\\ \hline
\multirow{2}*{$\sim$5\%}
& $K_1$ = 9 w/o CSinGAN &0.4730 & 0.6882\\
~ & $K_1$ = 9 w/ CSinGAN & \textbf{0.4920} & \textbf{0.7105} \\  \hline
\multirow{2}*{$\sim$7\%}
& $K_1$ = 12 w/o CSinGAN & 0.4765 &0.6829\\
~ & $K_1$ = 12 w/ CSinGAN & \textbf{0.4876}&\textbf{0.7029} \\ \hline
Annos & $K_1$ = 9 & AJI & Dice\\ \hline
\multirow{6}*{$\sim$5\%}
& $K_2$ = 3 w/o CSinGAN & 0.4574 & 0.6517 \\
~ & $K_2$ = 3 w/ CSinGAN & \textbf{0.4829} & \textbf{0.6920}\\ \cline{2-4}
~ & $K_2$ = 4 w/o CSinGAN & 0.4730 & 0.6882 \\
~ & $K_2$ = 4 w/ CSinGAN & \textbf{0.4920} & \textbf{0.7105} \\ \cline{2-4}
~ & $K_2$ = 8 w/o CSinGAN & 0.4340 & 0.6398\\
~ & $K_2$ = 8 w/ CSinGAN & \textbf{0.4586} & \textbf{0.6626}\\  \hline
\end{tabular}    
}
\caption{Hyper-parameters investigation of the proposed consistency-based patch selection on the TCGA-KUMAR dataset. Three values of $K_1$ (6, 9, 12) are tested when $K_2$ is set as 4. Three values of $K_2$ (3, 4, 8) are tested when $K_1$ is set as 9. The `Annos' donates the annotation proportion of the training set. Each CPS experiment is followed by CSinGAN or naive augmentation (w/o CSinGAN) to obtain the AJI and Dice values.}
\label{tab_param}
\end{table}

In TABLE~\ref{tab_param}, two  hyper-parameters $K_1$ and $K_2$ of the proposed CPS are studied on the TCGA-KUMAR dataset. `w/ CSinGAN' denotes the setting that uses a CSinGAN for data augmentation. `w/o CSinGAN' means that the CPS-selected samples are augmented by naive data augmentation instead of CSinGAN. All the experiment settings in TABLE~\ref{tab_param} adopt a Mask-RCNN as the nuclei segmentation model. For simplicity, pseudo labeling is not used in these experiments. Using CSinGAN for data augmentation, the AJI results range from 0.4517 to 0.4920 with $K_1$ as 6/9/12. It shows that the segmentation performance of our proposed method will improve when the $K_1$ is set from 6 to 9, but it does not improve when change it to 12. It may be due to that the newly added samples are of relatively low intra-patch consistency which could lead to low-quality CSinGAN-synthesized images. Thus, for the TCGA-KUMAR dataset, setting $K_1$ as 9 is slightly better. Setting $K_2$ as 4 is moderate and obtains higher AJI than other two candidate values.

\section{Conclusion}
In this paper, we first introduce a novel label-efficient framework that can segment nuclei with only 5\% pixel-level annotations, and presents even slightly better results than the fully-supervised baseline on some benchmarks. Second, we develop a consistency-based patch selection algorithm that leads to higher segmentation accuracy than K-means based or random selections. Furthermore, we propose a novel component-wise discriminator that considerably enhances the quality of image synthesis for a conditional single-image GAN. In short, our proposed framework reveals the importance of selecting which image pixels for labeling. The work points out a new trend in training medical image models with low-cost annotations.

\begin{bibliographystyle}{IEEEtran}
\begin{bibliography}{IEEEabrv,TMI}

\end{bibliography}
\end{bibliographystyle}
\end{document}